\documentclass[sigconf]{acmart}

\AtBeginDocument{%
  }

\copyrightyear{2025}
\acmYear{2025}
\setcopyright{rightsretained}
\acmConference[EICS Companion '25]{Companion Proceedings of the 17th ACM SIGCHI Symposium on Engineering Interactive Computing Systems}{June 23--27, 2025}{Trier, Germany}
\acmBooktitle{Companion Proceedings of the 17th ACM SIGCHI Symposium on Engineering Interactive Computing Systems (EICS Companion '25), June 23--27, 2025, Trier, Germany}\acmDOI{10.1145/3731406.3731972}
\acmISBN{979-8-4007-1866-3/2025/06}

\usepackage{subfig}
\usepackage{multirow}

\begin{document}

\title{InFL-UX: A Toolkit for Web-Based Interactive Federated Learning}

\author{Tim Maurer}
\email{s8timaur@stud.uni-saarland.de}
\orcid{0009-0004-5118-0440}
\affiliation{%
  \institution{Saarland University}
  \city{Saarbrücken}
  \country{Germany}
}

\author{Abdulrahman Mohamed Selim}
\email{abdulrahman.mohamed@dfki.de}
\orcid{0000-0002-4984-6686}
\affiliation{%
  \institution{German Research Center for Artificial Intelligence (DFKI)}
  \city{Saarbrücken}
  \country{Germany}
}
\author{Hasan Md Tusfiqur Alam}
\email{hasan_md_tusfiqur.alam@dfki.de}
\orcid{0000-0003-1479-7690}
\affiliation{%
  \institution{German Research Center for Artificial Intelligence (DFKI)}
  \city{Saarbrücken}
  \country{Germany}
}
\author{Matthias Eiletz}
\email{eiletz@proomea-group.com}
\orcid{0009-0009-4454-9808}
\affiliation{%
  \institution{ProOmea GmbH}
  \city{Wangen}
  \country{Germany}
}
\author{Michael Barz}
\email{michael.barz@dfki.de}
\orcid{0000-0001-6730-2466}
\affiliation{%
  \institution{German Research Center for Artificial Intelligence (DFKI)}
  \city{Saarbrücken}
  \country{Germany}
}
\affiliation{%
  \institution{University of Oldenburg}
  \city{Oldenburg}
  \country{Germany}
}
\author{Daniel Sonntag}
\email{daniel.sonntag@dfki.de}
\orcid{0000-0002-8857-8709}
\affiliation{%
  \institution{German Research Center for Artificial Intelligence (DFKI)}
  \city{Saarbrücken}
  \country{Germany}
}
\affiliation{%
  \institution{University of Oldenburg}
  \city{Oldenburg}
  \country{Germany}
}

\renewcommand{\shortauthors}{Maurer et al.}

\begin{abstract}
  This paper presents InFL-UX, an interactive, proof-of-concept browser-based Federated Learning (FL) toolkit designed to integrate user contributions into the machine learning (ML) workflow. InFL-UX enables users across multiple devices to upload datasets, define classes, and collaboratively train classification models directly in the browser using modern web technologies. Unlike traditional FL toolkits, which often focus on backend simulations, InFL-UX provides a simple user interface for researchers to explore how users interact with and contribute to FL systems in real-world, interactive settings. InFL-UX bridges the gap between FL and interactive ML by prioritising usability and decentralised model training, empowering non-technical users to actively participate in ML classification tasks. 
\end{abstract}



\maketitle

\section{Introduction and Related Work}

Traditional machine learning (ML) is often constrained by limited data, particularly in specialised domains where data acquisition is expensive or labour-intensive \cite{roh_survey_2021}. Moreover, technical barriers hinder direct input from domain experts, further delaying new data collection \cite{holstein_bridging_2024}. To overcome these issues, \citet{fails-olson_iml_2003} introduced interactive machine learning (IML), which enables non-technical users to train ML models by manually classifying data or correcting outputs. Unlike conventional ML, IML supports real-time updates based on user input, permitting focused, incremental refinements \cite{amershi_power_2014, dudley_review_2018, tusfiqur2022drg}. Extending this work, \citet{tseng_collaborative_2023} developed \textit{Co-ML}, a tablet-based application for collaboratively constructing ML image classification models that emphasises shared dataset design. In this paper, we build on these concepts by proposing a browser-based tool that facilitates collaborative IML using federated learning (FL).

ML models are traditionally trained on centralised datasets. However, in fields such as healthcare, data are distributed across multiple devices and cannot be shared due to privacy constraints. FL mitigates this issue by enabling decentralised training of a shared model while retaining data on client devices \cite{mcmahan_communication_2017}. FL aggregates local updates on a central server and comes in two main forms: synchronous and asynchronous. Synchronous FL \cite{mcmahan_communication_2017} requires all clients to train concurrently with the latest global model and submit updates together. In contrast, asynchronous FL \cite{xie_asynchronous_2019} allows clients to train and submit updates independently, with the server updating the global model upon each submission. Due to its flexibility, asynchronous FL is more suited to our application.
Recent advances in JavaScript-based deep learning frameworks, such as \textit{TensorFlow.js} \cite{smilkov_tensorflowjs_2019, ma_dl-webbrowser_2019}, have made browser-based deep learning (DL) feasible. For example, Google’s \textit{Teachable Machine} offers a no-code interface for local model training; however, it is limited to local training and does not incorporate FL. In contrast, browser-based FL frameworks proposed by \citet{lian_webfed_2022} and \citet{morell_browser-async-fl_2022} support FL but lack interactive elements, requiring users to supply data in predefined folders.

In this work, we introduce \textbf{InFL-UX}\footnote{The name is a combination of Interactive Machine Learning, Federated Learning, and User Experience.}, an interactive, browser-based FL toolkit that demonstrates a proof-of-concept (POC) implementation using contemporary web technologies. The application employs asynchronous FL via FedAsync \cite{xie_asynchronous_2019}, chosen for its simplicity and single-hyperparameter configuration, to enable collaborative model training. Users can upload datasets, define custom classes, and train models for various classification tasks. While much FL research relies on simulations, \textbf{InFL-UX} integrates FL with IML to prioritise user engagement and address FL challenges from a user-centric perspective. We aim to assist FL practitioners in embedding FL into intelligent user interfaces (UIs) and evaluate these setups through user studies. Future enhancements will incorporate advanced FL aggregation methods and additional ML tasks, such as data annotation, to broaden applicability.

\section{System Design}

\textbf{InFL-UX} utilises ONNX Runtime\footnote{\url{https://onnxruntime.ai/} (Accessed January 03, 2025)}, a cross-platform ML library that supports fast on-device inference and training in web browsers. It leverages modern browser APIs, including WebAssembly \cite{rossberg_webassembly_2022}—a low-level language offering near-native performance—and WebGPU \cite{ninomiya_webgpu_2024}, which facilitates high-performance GPU computations and supersedes WebGL\footnote{\url{https://www.khronos.org/webgl/} (Accessed January 03, 2025)}. This browser-based approach, which capitalises on the ubiquity of web browsers, ensures seamless compatibility with models from prevalent ML frameworks via the ONNX format. Developed in Python using Flask\footnote{\url{https://flask.palletsprojects.com/en/stable/} (Accessed January 03, 2025)}, the application integrates ONNX Runtime through JavaScript and employs Web Components\footnote{\url{https://www.webcomponents.org/} (Accessed January 03, 2025)} for UI modularity, thereby avoiding dependence on specific UI frameworks. Persistent storage is managed through IndexedDB \cite{bell_indexed_2024}, allowing user data, such as uploaded files and inference results, to be retained across sessions. The code for \textbf{InFL-UX} is publicly available at \url{https://github.com/tmaurer42/interactive-fl-poc}.

\begin{figure}
    \centering
    \includegraphics[width=\linewidth]{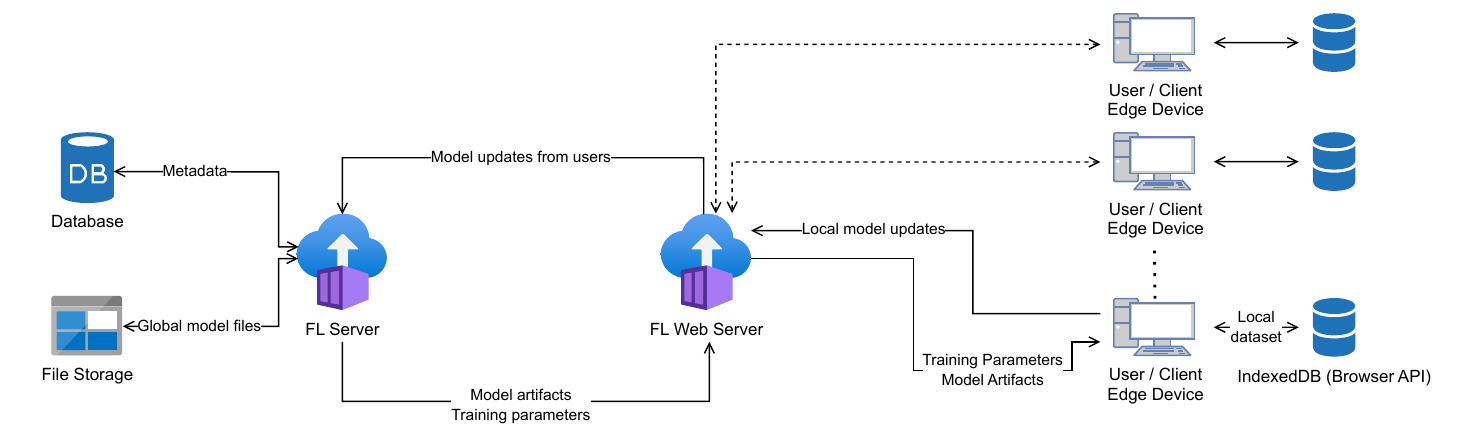}
    \caption{The system architecture overview.}
    \label{fig:architecture}
    \Description{This figure shows the overall architecture of the application, structured in a client-server setup.}
\end{figure}

\subsection{System and Application Design}

\begin{figure*}
  \centering
  \subfloat[Configuration of training parameters.\label{fig:configuration}]
      {\includegraphics[width=0.3\textwidth]{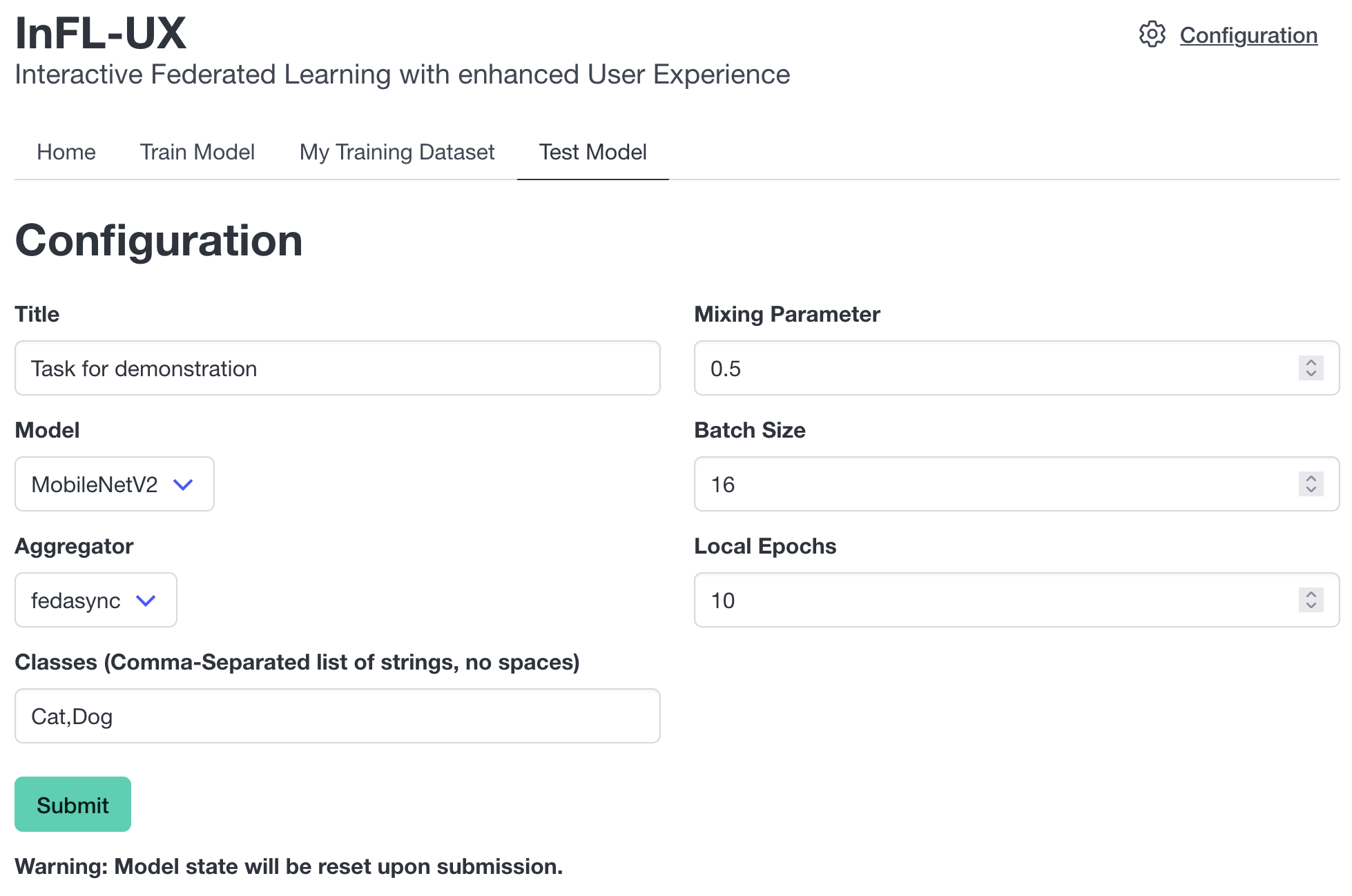}}
  \subfloat[Label review process for the training images; users can accept or correct suggested labels.\label{fig:review_process}]
      {\includegraphics[width=0.33\textwidth]{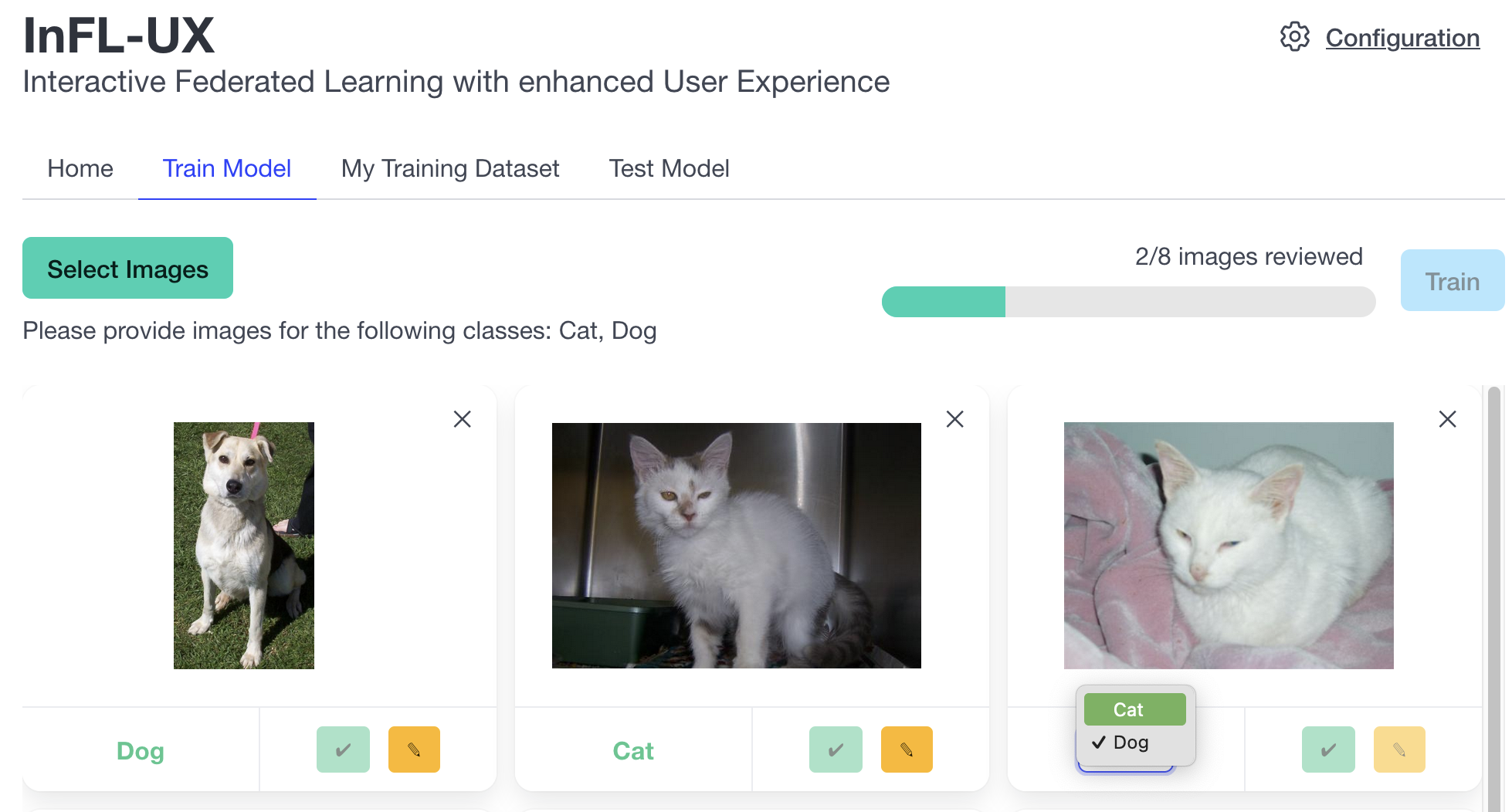}}
  \subfloat[Interface for testing the performance of the current model.\label{fig:model_testing}]
      {\includegraphics[width=0.33\textwidth]{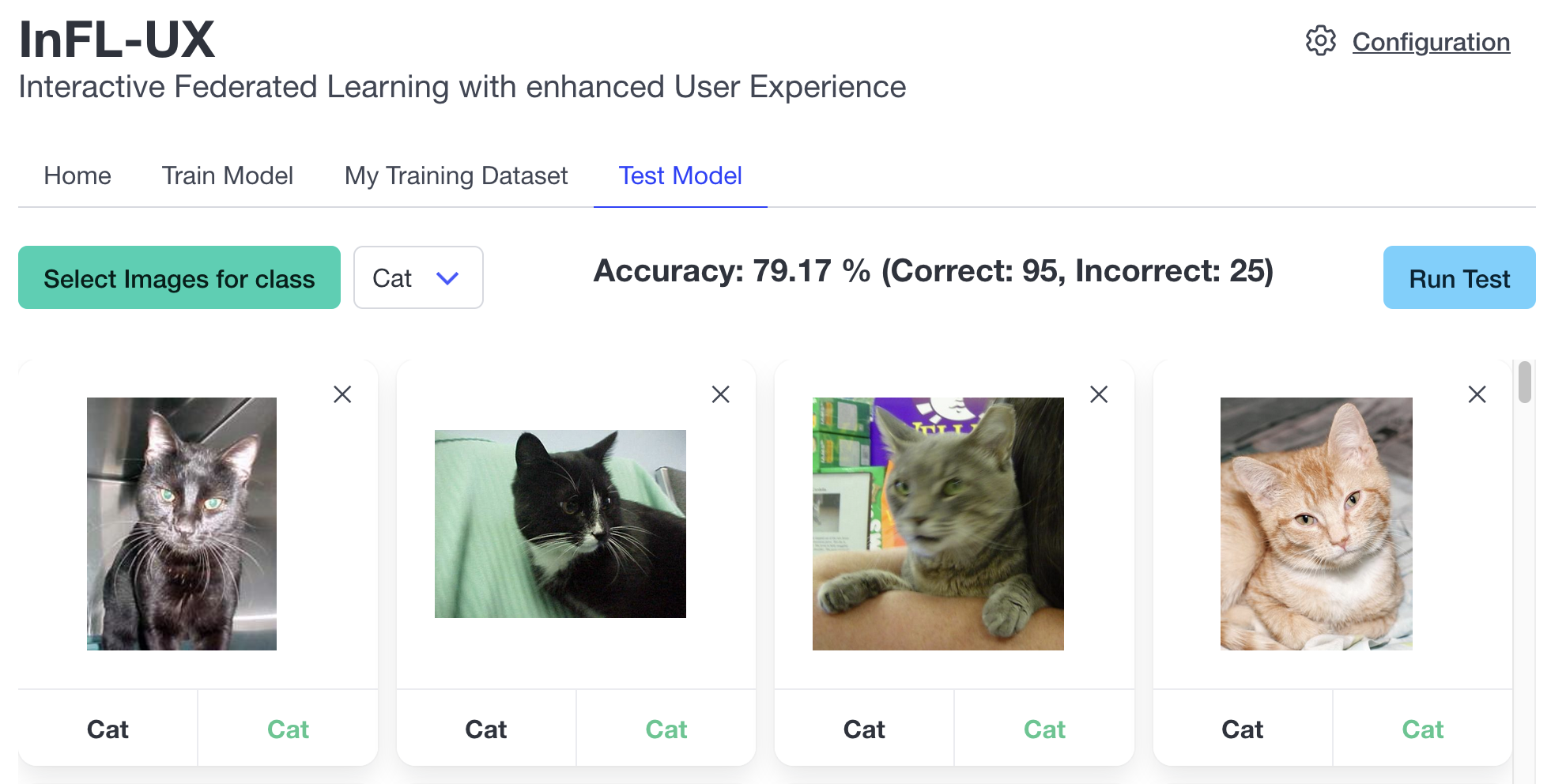}}
  \caption{Screenshots from InFL-UX showing (a) the configuration page, (b) the label review process, and (c) the model testing interface. The images used in the demo were taken from the Cats and Dogs Classification Dataset \textcopyright\ Bhavik Jikadara: \url{https://www.kaggle.com/datasets/bhavikjikadara/dog-and-cat-classification-dataset}, which is licensed under Apache 2.0.}
  \Description{This figure displays multiple screenshots from the InFL-UX application, highlighting various user interactions. Starting with the training configuration window in (a), then the label review process for the training images, which allows users to correct or accept the suggested labels in (b), and lastly, the interface for testing the trained model shown in (c).}
\end{figure*}

The architecture (Figure \ref{fig:architecture}) comprises two main components: the \textbf{FL Server} and the \textbf{FL Web Server}. The FL Web Server hosts the web application and delivers required files to client devices, while the FL Server manages client updates, maintains the global model, and provides the latest model version and training instructions on demand. This structure underpins the POC, which primarily addresses image classification; it can also accommodate tasks such as object detection and image segmentation through abstract classes and supports the simulation of multiple independent clients.
An administrator client configures the system via the configuration page (Figure \ref{fig:configuration}) by specifying the use case, selecting the aggregator and ML model, and setting training parameters. Clients then upload images, receive model-generated label suggestions, and may accept or adjust these labels (Figure \ref{fig:review_process}). Once reviewed, a training session is initiated locally, and upon completion, updates are automatically forwarded to the central server for aggregation. Clients can evaluate the global model using separate testing datasets (Figure \ref{fig:model_testing}).

Currently, the system implements a single FL aggregation strategy, i.e., FedAsync, and one ML model, i.e., MobileNetV2 \cite{sandler_mobilenetv2_2019}. Adding new aggregation strategies or ML models is straightforward, requiring only minimal code modifications: a new aggregation strategy is defined as a Python function with corresponding configurable parameters added to the administrator interface, and additional ML models are integrated via an expanded dropdown menu featuring extra PyTorch models\footnote{\url{https://pytorch.org/} (Accessed January 03, 2025)}. The system is deployed using Docker\footnote{\url{https://www.docker.com/} (Accessed January 03, 2025)}. Although designed for extensibility, \textbf{InFL-UX} does not support simulation tasks, which are handled by established frameworks such as Flower \cite{beutel_flower_2022}; instead, it focuses on validating aggregation methods in real-world settings.

\subsection{Limitations}

Several limitations emerged during development. The relatively new ONNX Runtime integration for the web offers incomplete training functionalities; for example, the optimiser’s learning rate is fixed at 0.001, and only a limited set of loss functions is available. Furthermore, while WebGPU currently supports inference, training is restricted to the WebAssembly backend. Browser-specific constraints also affect the application: the local dataset stored in IndexedDB is limited in size (varying by browser and available disk space), and the WebGPU API is presently available only in developer builds of modern browsers, which may be restricted by security policies. 

\section{Conclusion and Future Work}

\textbf{InFL-UX}, demonstrated the feasibility of an interactive, browser-based FL system that integrates state-of-the-art technologies from the UI to the server-side aggregation process. This approach allows embedding FL capabilities into IML systems, particularly in privacy-sensitive domains. Despite its limitations, the application effectively utilises ONNX Runtime and modern browser features (WebAssembly and WebGPU) to support client-side FL training, enabling collaborative training across devices while ensuring compatibility and encouraging user adoption.

Future work should involve user studies to assess real-world adoption, performance, and scalability and extend the application to additional computer vision tasks, such as image segmentation and object detection. Moreover, incorporating explainable AI techniques (e.g. class activation maps \cite{selvaraju_grad-cam_2017}, and concept-based modelling \cite{tusfiqur2025cbm}) could improve transparency and trust. Finally, enabling administrators to deploy new training tasks and allowing clients to participate in multiple tasks with distinct datasets would significantly enhance the system's utility.

\begin{acks}
This work was funded, in part, by the German Federal Ministry of Education and Research (BMBF) under grant number 01IW23002 (No-IDLE) and grant number 01IW24006 (NoIDLEChatGPT), by the Lower Saxony Ministry of Science and Culture (MWK) in the zukunft.niedersachsen program, and by the Endowed Chair of Applied AI at the University of Oldenburg.
\end{acks}

\bibliographystyle{ACM-Reference-Format}
\bibliography{main.bbl}

\end{document}